\documentclass[runningheads]{llncs}
\usepackage{graphbox}
\usepackage{amsmath}
\usepackage{multirow}
\usepackage{multicol}
\usepackage{graphicx}
\usepackage{arydshln}

\usepackage{todonotes}
\usepackage{algorithm}
\usepackage{algpseudocode}
\usepackage{booktabs}
\usepackage{soul}
\usepackage{hyperref}
\usepackage{cleveref}
\usepackage{balance}
\usepackage{subcaption}
\Crefname{equation}{Eq.}{Eqs.}
\Crefname{figure}{Fig.}{Figs.}
\Crefname{tabular}{Tab.}{Tabs.}
\Crefname{section}{Sec.}{Secs.}
\begin{document}
\title{Thinking Outside the Template with Modular GP-GOMEA}
%
%
\author{Joe Harrison\inst{1,2}\orcidID{0000-0002-3427-5251} \and
Tanja Alderliesten\inst{3}\orcidID{1111-0003-4261-7511} \and
Peter A.N. Bosman\inst{1,2}\orcidID{0000-0002-4186-6666}}
\authorrunning{Harrison et al.}
%
\institute{Centrum Wiskunde \& Informatica, Amsterdam, The Netherlands \and
Delft University of Technology, Delft, The Netherlands \and
Leiden University Medical Center, Leiden, The Netherlands}
\maketitle              

\begin{abstract}
The goal in Symbolic Regression (SR) is to discover expressions that accurately map input to output data. Because often the intent is to understand these expressions, there is a trade-off between accuracy and the interpretability of expressions. GP-GOMEA excels at producing small SR expressions (increasing the potential for interpretability) with high accuracy, but requires a fixed tree template, which limits the types of expressions that can be evolved. This paper presents a modular representation for GP-GOMEA that allows multiple trees to be evolved simultaneously that can be used as (functional) subexpressions. While each tree individually is constrained to a (small) fixed tree template, the final expression, if expanded, can exhibit a much larger structure. Furthermore, the use of subexpressions decomposes the original regression problem and opens the possibility for enhanced interpretability through the piece-wise understanding of small subexpressions. We compare the performance of GP-GOMEA with and without modular templates on a variety of datasets. We find that our proposed approach generally outperforms single-template GP-GOMEA and can moreover uncover ground-truth expressions underlying synthetic datasets with modular subexpressions at a faster rate than GP-GOMEA without modular subexpressions.
\keywords{GOMEA\and Symbolic Regression\and Genetic Programming\and Explainable AI\and Automatically Defined Functions}
\end{abstract}
\section{Introduction}
Explainable AI (XAI) is becoming increasingly important as decisions in critical domains, such as healthcare, finance, and criminal justice, are made more frequently with the support of machine learning algorithms \cite{doshi2017towards}. By clarifying the often opaque decision-making processes of these models, XAI not only enhances fairness by revealing potential biases and offering opportunities for mitigation but also builds trust among end-users who rely on these systems \cite{lipton2018mythos}. Trust is important for the adoption of XAI, as \cite{ribeiro2016should} highlight that users are unlikely to use a model they do not trust. Moreover, with regulatory frameworks like the European Union's General Data Protection Regulation mandating the "right to explanation", XAI is not just a desire but is becoming a legal necessity \cite{goodman2017european,samek2017explainable}.

Symbolic Regression (SR) is a form of XAI where the relationship between input variables and output predictions within a given dataset is modelled as a symbolic expression. Unlike typical regression methods, SR concerns optimizing both the structure and parameters of a symbolic expression \cite{kommenda2020parameter,schmidt2009distilling}. The goal is to find symbolic expressions that achieve high predictive accuracy without unnecessary complexity, so as to promote interpretability. The interpretability of an expression is typically influenced by the size of its structure, in particular, the number of operators included \cite{virgolin2021improving,kommenda2018local,mei2022explainable}. While neural networks often excel in accuracy, their large-scale structure compromises the direct interpretation of trained models without using post-hoc explanation methods. Moreover, post-hoc explanations are often not exact or complete. Conversely, expressions resulting from SR tend to be smaller, yet they may sacrifice some accuracy in exchange for enhanced interpretability in the form of algorithmic transparency, simulatability or traceability, and, in the case of Genetic Programming (GP), decomposability - three key factors in XAI according to \cite{lipton2018mythos}.

One GP method that is known to be efficient and effective at generating compact models, including for SR, is the Gene-pool Optimal Mixing Evolutionary Algorithm for Genetic Programming (GP-GOMEA) \cite{virgolin2017scalable,virgolin2021improving}. GP-GOMEA draws its strength from the integration of probabilistic model-building techniques from Estimation of Distribution Algorithms and local search principles. Among all algorithms tested on the SR benchmark known as SRBench~\cite{la2021contemporary}, GP-GOMEA resides on the knee of the non-dominated front, effectively balancing accuracy and expression size. 

A fixed tree template is used in GP-GOMEA to represent an expression. Often, this represents a full tree of limited depth. While this partially the reason why GP-GOMEA is so efficient and effective at learning compact models, this rigidity also poses challenges. Specifically, various (regression) problems may demand expression structures that do not fit the predefined template (e.g., a deep unbalanced tree may be needed, while the need for this structure may not be known a priori and can then thus not be pre-configured).

Another limitation, which is common to most forms of GP, is that a single expression is searched for that is defined directly in terms of the input features, coefficients, and operators. In other words, there is no mechanism for detecting and exploiting the potential re-use of parts of the expression, or, more generally: modularity. In SR, however, it may sometimes be possible to hierarchically decompose expressions into smaller subexpressions and re-use certain subexpressions. Decomposability is an important factor in the interpretability of models \cite{lipton2018mythos}. Moreover, the re-use of subexpressions arguably can lower the total effort spent on interpreting full expressions \cite{koza1993hierarchical}, allowing for the representation of larger expressions, offering potentially better learning capacity (i.e., the ability to obtain higher accuracy), without compromising interpretability much.

As suggested in \cite{virgolin2021improving} GP-GOMEA can also potentially benefit from such repeating subexpressions. A first step in this direction was the work presented in \cite{harrison2022gene}, which introduced Cartesian Genetic Programming (CGP) to GOMEA (CGP-GOMEA). The CGP representation enabled the automatic discovery of repeating subtrees with GOMEA (i.e. re-use of duplicate subexpressions with the same input arguments). However, due to the complexity involved with the CGP representation, CGP-GOMEA has to be constrained with a rather restrictive template. While this template provides more flexibility than GP-GOMEA in terms of re-use, it still limits the overall flexibility and expressiveness of the models it generates. Moreover, the search effort required to find good expressions also increases. Embedded CGP, as described in \cite{walker2008automatic}, extends the capabilities of the CGP representation by allowing the discovery of repeating subexpressions that can take different inputs and serve as general functions. These functions, coined by Koza as Automatically Defined Functions (ADFs), include both duplicate and functional subexpressions. While embedded CGP introduces this valuable function-like behaviour, it still does not support the hierarchical calling of subexpressions within other subexpressions, which limits its modularity. Hierarchical Automatically Defined Functions (HADFs) \cite{koza1993hierarchical,koza1994scalable} are an extension of ADFs where other GP-trees are included in both the main GP-tree's operator set as subexpressions and the operator sets of other subexpressions, allowing for the hierarchical calling of subexpressions. A more detailed overview of modularity in genetic programming can be found in \cite{gerules2016survey}.

In this paper, we introduce \textbf{Modular GP-GOMEA}, a novel variant of GP-GOMEA that enables expressions to be found that are beyond a typical single full-tree template, by explicitly encoding and \textbf{jointly evolving subexpressions} that can be hierarchically used. The approach proposed in this paper is similar to Koza's HADFs, but differs in a few key ways: building on and leveraging the potential of GP-GOMEA, small fixed-sized templates are used for the subexpressions, -rather than arbitrarily large GP-trees, which arguably enhances their interpretability. Additionally, the terminal sets of these subexpressions include not only function input arguments but also input features and coefficients, allowing for more expressive subexpressions. The importance of this more extensive terminal set can be seen in Appendix \ref{sec:inputs}. 

\section{Methods}
We divide our method explanation into two parts. First, we explain how GP-GOMEA works in general. Subsequently, we describe how multiple subexpressions are included in the representation and how variation is adjusted for this in the new version of GP-GOMEA that we call \textbf{Modular GP-GOMEA}. 
\subsection{GP-GOMEA}
\begin{figure}
    \centering
    \includegraphics[width=\textwidth]{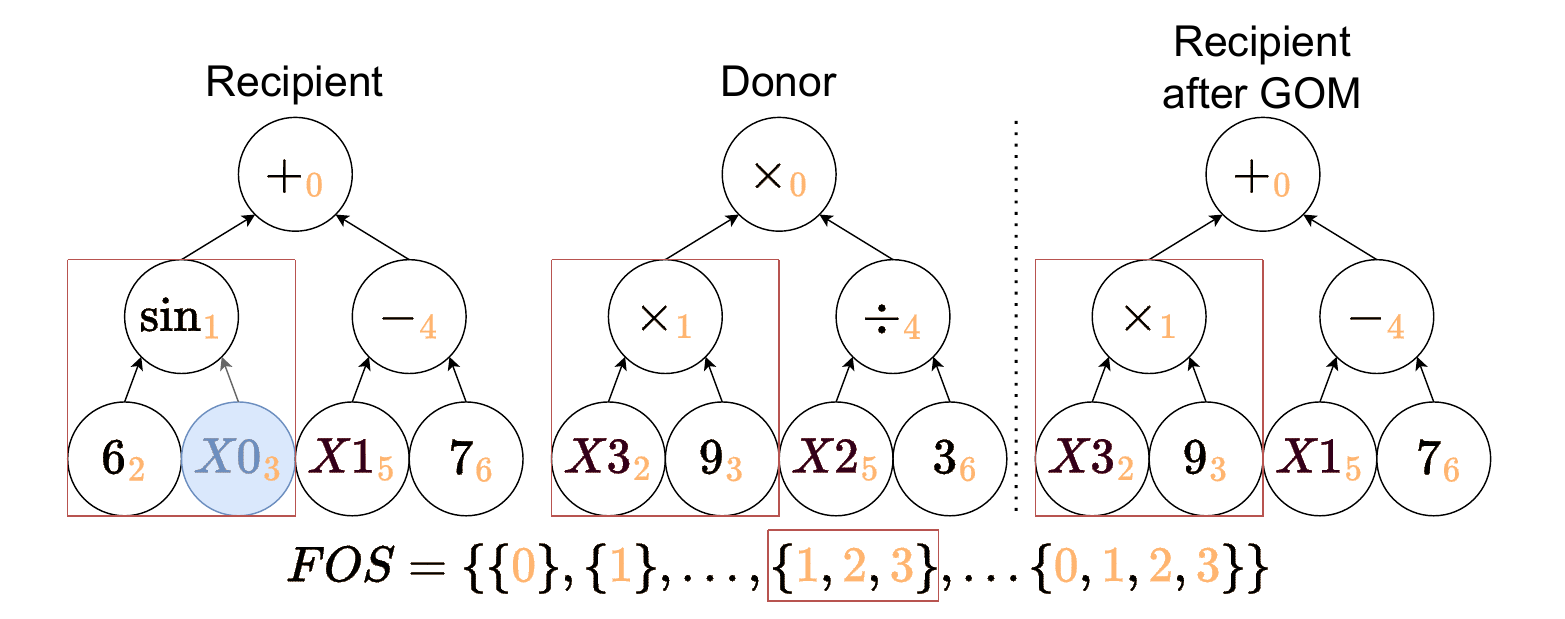}
    \caption{Example of an FOS and a GOM swap. The red rectangle indicates an FOS subset that undergoes GOM. Node indices (in pre-order) are indicated with a subscript.}
    \label{fig:GOMEA_explanation}
\end{figure}
In GP-GOMEA, the fixed-size tree templates that the symbolic expressions that can be found, adhere to, are represented as strings. Each string index corresponds to a fixed position in the tree template. Specifically, pre-order tree traversal is used for this (see Figure \ref{fig:GOMEA_explanation}). Some loci have a strong dependence on each other, e.g., it could be that locus 4 in the donor may only be beneficial to the individual if it is used in conjunction with having $x_1$ and $9$ at loci 5 and 6, respectively. For efficient and effective search, it is beneficial to group loci with strong dependence together and swap them simultaneously \cite{thierens2011optimal}.

GP-GOMEA uses linkage information within the population to group string indices into multiple sets comprising a Family Of Subsets (FOS). Often, a linkage tree FOS is used, which is a hierarchically structured set of sets. Conceptually, this set can be constructed by initially placing every string index in a singleton set and then iteratively merging two sets that exhibit the strongest dependence (also called linkage strength) until all indices are in one set. In non-modular GP-GOMEA the subset containing all indices, i.e. corresponding to the whole tree, is removed from the FOS to avoid swapping whole trees (i.e. the FOS size is $2l - 2$, where 
 $l$ is the length of the string). Mutual Information (MI) estimated from the population using maximum likelihood estimates, is used in this paper as a proxy for assessing linkage strength between indices. To approximate linkage between higher-order groupings, these indices are clustered hierarchically using the Unweighted Pair Group Method with Arithmetic Mean\cite{gronau2007optimal} (UPGMA) with the MI matrix, resulting in the construction of the linkage tree FOS. The MI matrix of the initial population should not contain any linkage. However, because not every terminal or non-terminal can be placed in every string position (e.g., non-terminals cannot be placed in the leaves of the  template), when generating trees using concical tree initialization methods in GP, the initial MI matrix does not automatically have this property. As in \cite{virgolin2021improving},  to remove this bias, the MI matrix of the initial population is therefore kept and subtracted from MI matrices that are estimated in subsequent generations.

Variation in GP-GOMEA is accomplished through Gene-pool Optimal Mixing (GOM), a process that is applied to a clone of each individual in the population in every generation. During GOM, each subset in the FOS is considered in a randomized order, and the parts of the recipient individual identified by the loci in the subset are replaced by the homologous parts of a donor solution randomly chosen from the population (see e.g. the swap of FOS subset $\{1,2,3\}$ in Fig. \ref{fig:GOMEA_explanation}). If any of the loci in the subset are connected to the output (i.e. non-introns) and at least one locus changes meaningfully (e.g. from $+$ to $\times$), the fitness of the changed individual is calculated. If the fitness has improved or remains equal, the change is accepted in a hill-climbing manner. 

Coefficients can be any floating point value. However, if each unique coefficient is given its own index, then the MI-matrix cannot be properly computed because mutual information relies on repeated occurrences to estimate statistical dependencies. In classical GP-GOMEA, all coefficient values are sampled at initialization and then left unchanged, which results in an extremely large set of terminals. To address this and to capture dependencies effectively, we use online binning to group similar coefficients into a fixed number of bins. In \cite{virgolin2021improving}, online binning is used to bin coefficients per tree template index; similarly, we use 25 coefficient bins as in \cite{schlender2024improving}. After each mixing operation in GOM, we mutate each coefficient in the individual as done in \cite{virgolin2022coefficient}. If the individual improves in terms of training error the changes are kept and are otherwise reverted. The coefficient mutation rate is set to $1.0$ and the coefficient mutation step size to $0.1$. The step size is decreased by a factor of 10 every 5 generations no change has been made to the elitist archive.

First, a random population of binary-unary trees is generated using the half-and-half method (50\% 'grow', 50\% 'full') \cite{koza1994genetic}. The probability of sampling a terminal node (input feature or coefficient) is set to 50\% for the 'grow' method. The probability of sampling a coefficient, given that a terminal node is sampled is set to 50\%, this differs from the implementation of GP-GOMEA in \cite{virgolin2021improving} where the probability of sampling a coefficient is $\frac{1}{1+\text{\#input features}}$.  Because GP-GOMEA operates on fixed-sized templates, the unused loci in the template that can occur using the grow method are filled up with random intron nodes which are sampled randomly from the operator set. When a unary operator such as $sin_1$ in the donor of Fig. \ref{fig:GOMEA_explanation} has two incoming children operators it will take the leftmost operator as input (coefficient $13_2$).

A multi-objective elitist archive is maintained and updated after each improvement. This enables end-users to have more expressions to choose from at the end of a run. In this paper, the archive keeps track of individuals that are non-dominated in terms of two objectives: expression size and $R^2$, using an adaptively gridded archive from \cite{luong2012elitist} with a target size of 100 solutions. Although expression size is not explicitly optimised, the solutions found whilst improving the $R^2$ still form a reasonable trade-off front, in part because during initialization also small expressions are generated already.

The mechanics of GP-GOMEA as used in this work are largely the same as in \cite{virgolin2021improving}. We have listed key differences below:
\begin{itemize}
    \item 50\% probability of sampling a coefficient as terminal during initialization
    \item 25 bins for grouping coefficients to indices
    \item Coefficient mutation occurs during GOM as in \cite{virgolin2022coefficient}
\end{itemize}
These differences lead to a small but insignificant performance gain in 3/5 of the tested datasets. We also want to point the interested reader to \cite{virgolin2021improving} for a more detailed explanation of GP-GOMEA. 

\subsection{Modular GP-GOMEA}
\label{sec:modular}
\begin{figure}
    \centering
    \includegraphics[width=\textwidth]{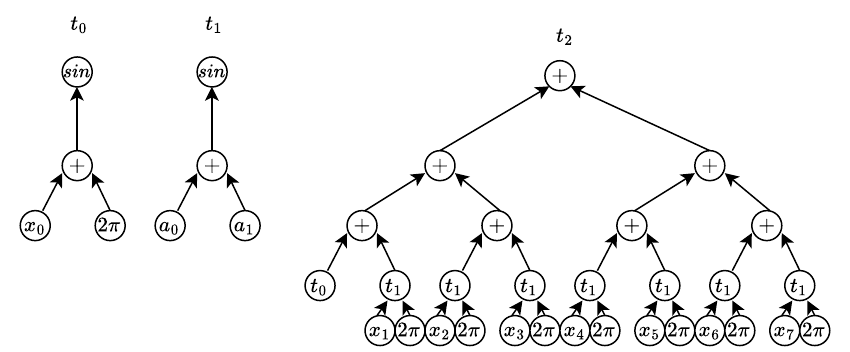}
    \caption{Example of an individual with three trees $t_0$, $t_1$, and $t_2$. $t_2$ is the output of the tree. $t_1$ is used as a function by $t_2$. The \textit{argument nodes} in a tree $i$ are indicated by $a_n$ and point to the arguments in the tree $j$ ($j>i$) that calls the tree $i$ (i.e, as a function), e.g. $a_0$ points to $x_1$ for the leftmost call to tree (function) $t_1$ in tree $t_2$.}
    \label{fig:example_tree}
\end{figure}

We use a multi-tree formulation in which an individual comprises not just one tree template, but a vector of $n$ tree templates (see Figure \ref{fig:example_tree}). Unlike the typical multi-tree setting where each tree serves as a separate output, in this paper only the last tree ($t_{n-1}$) functions as the output of the individual. Any tree at location $i$ in the representation can make use of any tree $j$ with $j<i$. We impose this restriction to prevent cyclic references that may lead to infinite loops. For example, in Figure \ref{fig:example_tree}, the final tree, $t_2$, can use $t_1$, but not vice versa. Note that it is possible that all nodes in a tree $t_j$ with $j<n-1$ are not connected to the output and are thus introns.

\textbf{Disambiguation notice} - The term subexpression in other works can refer to the expression represented by a part of a tree (i.e., a subtree). However, in this work, 'subexpression' specifically refers to any tree in the multi-tree representation that the last tree can reference.

In addition to a function set, coefficients, and input features used in GP-GOMEA\cite{virgolin2021improving}, we introduce 2 special operators. \textbf{1.} the \textit{subexpression node}, a function node that can call a preceding tree as a function and use its output, \textbf{2.} the \textit{argument node}, a terminal node that points to an argument of the tree calling it according to its associated index (see Figure \ref{fig:example_tree}). In this paper, we limit the maximum arity of all tree templates to 2. \textit{Subexpression nodes} can therefore, only have 2 arguments and the \textit{argument nodes}, can only have 2 possible indices. During initialization, additional subexpressions nodes $i-1,i-2...,0$ can be sampled for tree $i$. All trees, except tree 0, have additional argument nodes available for sampling. Subexpressions can be used as a function (see $t_1$ in Fig \ref{fig:example_tree}) or as a terminal of the tree (see $t_0$ in Fig \ref{fig:example_tree}) where the terminal nodes remain the same.

\begin{algorithm}
\caption{FOS Learning and Contraction for Multi-Trees}
\begin{algorithmic}[1]
\State \textbf{Input:} Population of multi-trees with $n$ trees per individual
\State Initialize $\text{FOSes} = [\ ]$ 
\For{each tree index $i = 0$ to $n-1$} 
    \State Learn $\text{FOS}_i$ from the $i$-th trees across the population
    \If{$i$ is the index of the output tree (last tree)}
        \State Remove the subset from $\text{FOS}_i$ that contains all nodes of the tree
    \EndIf
    \State $\text{FOSes}[i] \gets \text{FOS}_i$
\EndFor
\State \textbf{Flattened FOS:}
\State $\text{Flattened FOS} \gets \bigcup_{i=0}^{n-1} \left\{ F_i \mid F_i \in \text{FOS}_i \text{, each element tagged with } i \right\}$
\State \textbf{Output:} Flattened FOS
\end{algorithmic}
\end{algorithm}

Each generation a separate FOS is learned for each tree within the multi-tree. Only the FOS of the last tree lacks a subset containing all nodes in the tree to prevent swapping the entire output tree. Conversely, the FOS of the other trees within the multi-tree have a subset with all nodes, allowing for the swapping of entire subexpressions. Before performing GOM the separate FOSes are contracted into one large FOS, which is then subsequently shuffled. This is done to prevent any bias that may have unintended negative side-effects due to the fact that the trees in the multi-tree representation are interdependent as some trees may call other trees.

\begin{algorithm}
\caption{Performing GOM for Multi-Trees}
\begin{algorithmic}[1]
\For{each individual in the population}
    \State Clone the individual into offspring
    \State Shuffle the Large FOS with indices
    \For{each $F_i$ in the shuffled Flattened FOS}
        \State Sample a donor individual from the population
        \State Copy operators corresponding to $F_i$ from the donor to the offspring
        \State Check if the swap leads to meaningful change 
        \If{meaningful change has occurred}
            \State Evaluate the fitness of the offspring
            \If{the fitness of the offspring has not improved}
                \State Revert the changes (restore original operators and fitness)
            \Else 
            \State Update MO-archive with the offspring
               \EndIf
        \EndIf
    \EndFor
    \State add offspring to offspring population
\EndFor
\State offspring population replaces population
\end{algorithmic}
\end{algorithm}

\section{Experiments and Results}
In this section, we first describe the general setup of the experiments. We then create 5 new synthetic datasets based on ground-truth expressions in which subexpressions are used. We then perform experiments on these synthetic datasets to get insights into the ability to recover ground-truth expressions and study the optimal population size of Modular GP-GOMEA. Using the best population size determined in the previous experiment we perform new experiments to show the relation between depth, number of multi-trees, and accuracy. We then give insights into the use and re-use of subexpressions. Finally, we provide qualitative insights into one of the subexpressions evolved by Modular GP-GOMEA.

\subsection{General Setup}
We run experiments for 5 real-world datasets (see Table~\ref{tab:realworld}) and 5 synthetic datasets (see Section~\ref{sec:synth}). 

\begin{table}[h]
\tabcolsep=2mm
\centering
\begin{tabular}{ccccc}
\toprule
Dataset & \#Samples & \#Features & Mean target & Variance target \\
\midrule
Airfoil    & 1503      & 5          & 124.8    & 6.9          \\
Bike Daily     & 731       & 11         & 4504.3     & 1935.9          \\
Concrete   & 1030      & 8          & 35.8     & 16.7         \\
Dow Chemical & 1066      & 57         & 3.0    & 0.1         \\
Tower      & 4999      & 25         & 342.1    & 87.8         \\
\bottomrule
\end{tabular}
\caption{Summarizing information on real-world datasets. Two columns that are subtotals of the target are removed from the Bike Daily dataset to avoid trivial expressions.}
\label{tab:realworld}
\vspace*{-6mm}
\end{table}

Each run of (Modular) GP-GOMEA is performed using a separate core of an AMD EPYC ROME 7282. A run is terminated either when a time budget of 6 hours is reached, or when the elitist archive has not changed for 100 consecutive generations. To compare performance we use the coefficient of determination ($R^2$). 

In Table~\ref{tab:general_experiment_info}, we provide an overview of the general settings that we used in all of the experiments in this paper.

\begin{table}[htbp]
\tabcolsep=2mm
\centering
\begin{tabular}{ll}
\toprule
Parameter                        & Setting                  \\
\midrule
Coefficient sampling             & $\sim U(min_{\text{target}},max_{ \text{target}})$         \\
Probability sampling coefficient & 50\%                     \\
Operator set                   & $+,-,*,/,sin,cos,log,sqrt,\text{input features},$\\&$\text{subexpression}_t^*,\text{argument}_0^\dagger,\text{argument}_1^*$ \\
\# Repetitions                   & 30                       \\
Stopping criterion               & 6  hours, 100 consecutive non-improvements                  \\
Max batch size                       & 2048                     \\
\bottomrule
\end{tabular}

\caption{General experiment settings, unless indicated otherwise. \textbf{$\dagger$} Arguments and subexpressions in operator set according to Section \ref{sec:modular}. Coefficients are sampled uniformly between the smallest and largest values of the target in the training set.}
\label{tab:general_experiment_info}
\vspace*{-6mm}
\end{table}

\subsection{Synthetic Datasets}
\label{sec:synth}
It is unknown whether the real-world regression datasets in Table \ref{tab:realworld} are decomposable in any way or, more specifically, contain any re-usable subexpressions because the ground-truth expression is unknown. However, as the design of modular GP-GOMEA is for a large part aimed at recovering expressions with re-used subexpressions, we construct 5 synthetic datasets for which we know the ground-truth expression and its decomposition into subexpressions.

We choose a maximum expression depth of 7 such that each configuration of GP-GOMEA in theory can recover the ground-truth expression. Each ground-truth expression has a unique characteristic which we outline below. To gain insights into the capabilities of modular GP-GOMEA we create a synthetic dataset with 1000 samples for each subexpression. The expressions pertaining to our synthetic datasets are the following:

\begin{enumerate}
    \item $\sum_{i=1}^8 \sin(x_i + x_0)$ \\
    8 times the same subexpression.
    \item $\sin(x_2 \times x_3) + \sum_{i=1}^7 \sin(x_i \times x_0)$ \\
    7 times the same subexpression with one different subexpression.
    \item $\sum_{i=1}^4 \sqrt{|\sin(x_i \times x_0)|}$ \\
    4 times the same, but more complex (compared to e.g. the subexpression in expression 1), subexpression.
    \item
    $f_0(f_1(x_0, x_1), f_1(x_2, x_3)) + f_1(f_0(x_0, x_1), f_0(x_2, x_3))$ \\
    with $f_0(a,b)=sin(a + b), f_1(a,b)=cos(a \times b)$ \\
    Subexpressions that are used within another subexpression and vice versa.
    \item $f_0(a,b,c)=\cos(a * \sin(\frac{b}{c}))$ \\
    $f_0(x_0,x_1,x_2) + f_0(x_0,x_2,x_1) + f_0(x_1,x_0,x_2) + f_0(x_1,x_2,x_0)$ \\
    Subexpression with an arity of 3 and permuted inputs. 
\end{enumerate}

Where the input feature $x_i$ is the $i + 1$th prime number (2,3,5,7,11,13,17,19) multiplied by a random number sampled from $\mathcal{U}[0,1]$.
\subsection{Finding the right population size}
We first perform an experiment to find an appropriate population size for the two different variants of GP-GOMEA by comparing the average $R^2$ obtained using population sizes 1024, 2048, 4096, 8192, 16384 and 32768 on each of the 5 synthetic datasets. The modular configuration uses 4 tree templates of depth 4 (4x4), and the non-modular configuration a single tree template of depth 7 (7x1). The tree template depths and the number of tree templates in an individual are chosen so that enough structure is available to construct any of the synthetic expressions above. Note that the 4x4 configuration, for example for expression 1, must use subexpressions to reconstruct the expression. More experiments regarding the depth and number of multi-trees for the modular setting are performed in section \ref{sec:varheight}.

\begin{algorithm}
\caption{Template-constrained GP}
\begin{algorithmic}[1]
\State \textbf{Input:} Population of individuals
\State \textbf{Output:} New population after crossover, mutation, and selection
\State Initialize new population $\gets [\ ]$
\For{each individual in the population}
    \State Choose a random donor individual
    \State Perform uniform crossover on multi-trees between the donor and individual to create a new individual
    \For{each multi-tree in the new individual}
        \State Sample a random subtree from another random donor at the same depth or lower within the corresponding multi-tree
        \State Transplant the donor subtree into the new individual
    \EndFor
    \State Apply point mutation
    \State new population $\gets$ new population $\cup$ new individual
\EndFor
\State New population $\gets$ tournament selection(population $\cup$ new population)
\end{algorithmic}
\end{algorithm}

Additionally, we compare two GP configurations (GP 4x4 and GP 7x1) in which the largest population size (32768) is used. Smaller population sizes result in worse $R^2$ for GP and are omitted for clarity. The GP configurations are restricted by the same maximum tree size as the templates used in  GP-GOMEA because we are interested in finding potentially interpretable (sub)expressions of a manageable size. In our template-constrained GP, each individual is modified independently to produce one offspring. Crossover is applied in two stages. First, uniform crossover is performed at the level of subtrees: for each tree in the multi-tree representation, there is a 50\% chance it is replaced by the corresponding tree from a randomly selected donor individual. Second, for each tree (whether swapped or not), a subtree crossover is applied: a random subtree is sampled from the individual’s tree, and a donor subtree (from a separately sampled donor individual) is selected at the same depth or lower to ensure a valid tree structure. The donor subtree is then transplanted. Mutation is done through point mutation on each node within a sampled mask of size $1+\sqrt{\text{\#nodes}}\times |x|$, where $x \sim \mathcal{N}(0,1)$. Selection is performed through tournament selection on the parents + offspring with tournament size 4.
\begin{figure}[H]
    \centering
    \includegraphics[width=\textwidth]{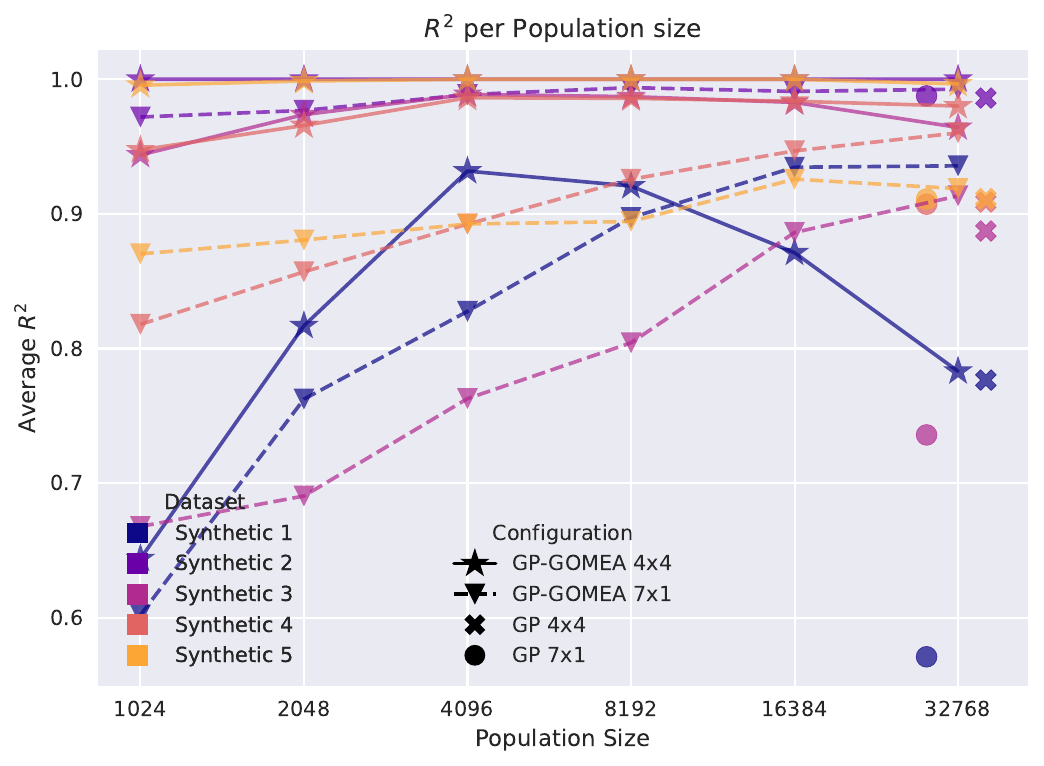}
    \caption{Comparison of average $R^2$ between different configurations of GP-GOMEA (4 trees of depth 4, 1 tree of depth 7) and depth-constrained GP. Positions of the markers of the GP configurations are shifted sideways for clarity.}
    \label{fig:r2perpopsize}
\end{figure}
In Figure \ref{fig:r2perpopsize} we observe that the average $R^2$ of the non-modular GP-GOMEA configuration has a general upward trend that increases with the population size within the population size range that we experimented with. The accuracy of the MI-matrix estimation increases with population size \cite{virgolin2021improving} leading to a more salient FOS being learned. Furthermore, non-modular GP-GOMEA must have the right building blocks for the same subtree at multiple node locations since it cannot re-use subtrees in a different location, whereas the modular configuration in theory only needs to evolve a subexpression once and can then re-use it in multiple node locations by using the subexpression node. 

Except for synthetic dataset 1 with a population of 16k and 32k, the performance of the non-modular configuration overall is worse than the modular configuration, which in some cases even reaches higher average $R^2$ scores with smaller population sizes. The GP formulation performs even worse, even though this configuration uses mutation, meaning that the initial population does not necessarily need to contain the right building blocks in the initial population. Interestingly, the 4x4 GP configuration, similar to how ADFs in \cite{koza1994genetic} outperform their non-modular counterpart, outperforms its 7x1 counterpart in 4/5 datasets, suggesting the modular representation can increase performance across optimization methods.

\begin{figure}[H]
    \centering
    \includegraphics[width=\textwidth]{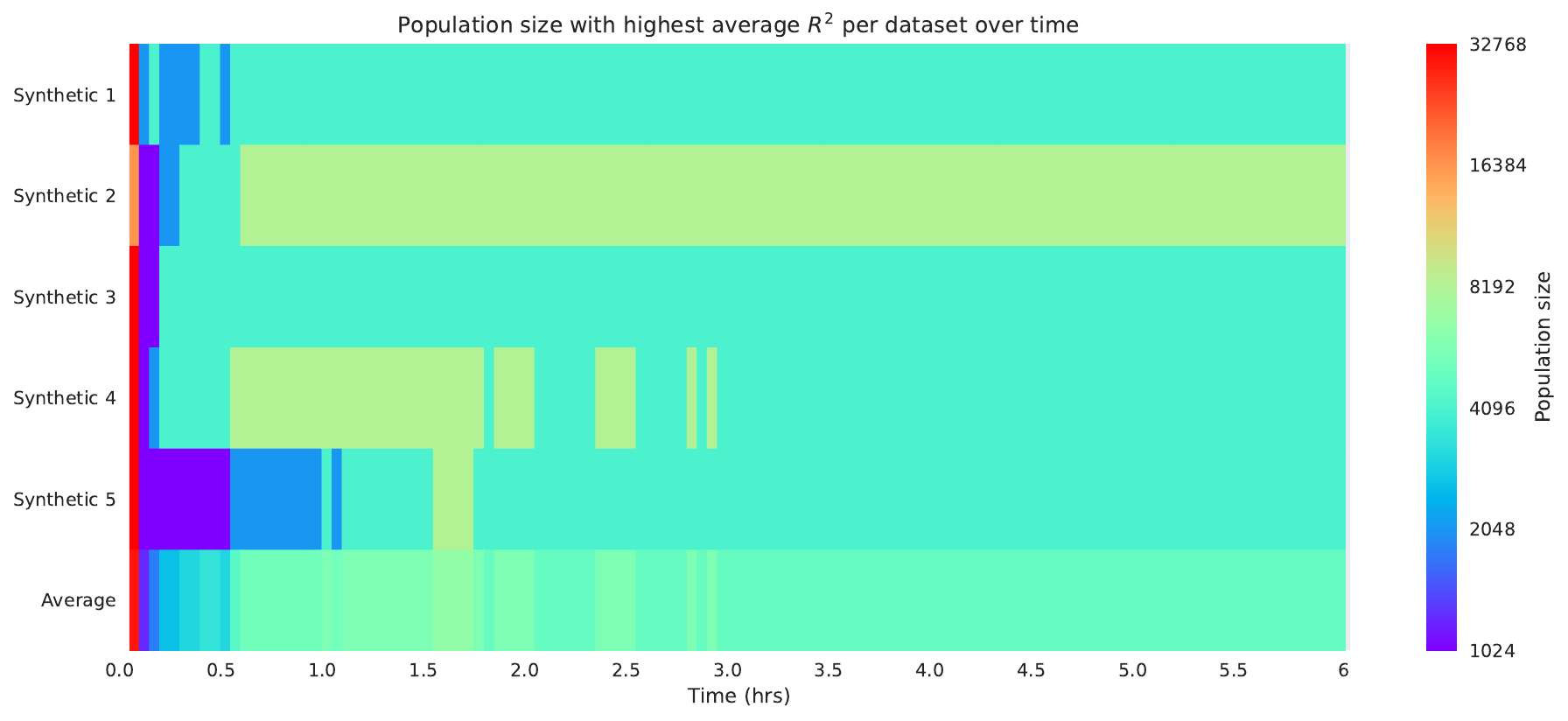}
    \caption{The population size with the highest average $R^2$ overall runs over time for the 4x4 modular configuration of GP-GOMEA. $R^2$ is interpolated linearly to obtain specific time points. The overall average is calculated as the average population size over the highest averages. At the 1-hour mark, the average population size is 5324.}
    \label{fig:optimalpopsizepertime}
\end{figure}

The best-tried population size within the time budget for the non-modular configuration of GP-GOMEA is the largest. In the modular configuration of GP-GOMEA, the average best-tried population size is 8600. Too small of a population leads to worse $R^2$ due to the right building blocks being unavailable in the initial population. Too large of a population leads to a decline in $R^2$ due to the algorithm reaching fewer generations within the time budget because larger population sizes require more evaluations per generation. When further restricting the time budget to 1 hour, as done in the following experiments in subsection \ref{sec:varheight}, the best tried average population size for the modular configuration lies closest to 4096 (see Figure \ref{fig:optimalpopsizepertime}).

\begin{figure}[h]
    \centering
    \includegraphics[width=\textwidth]{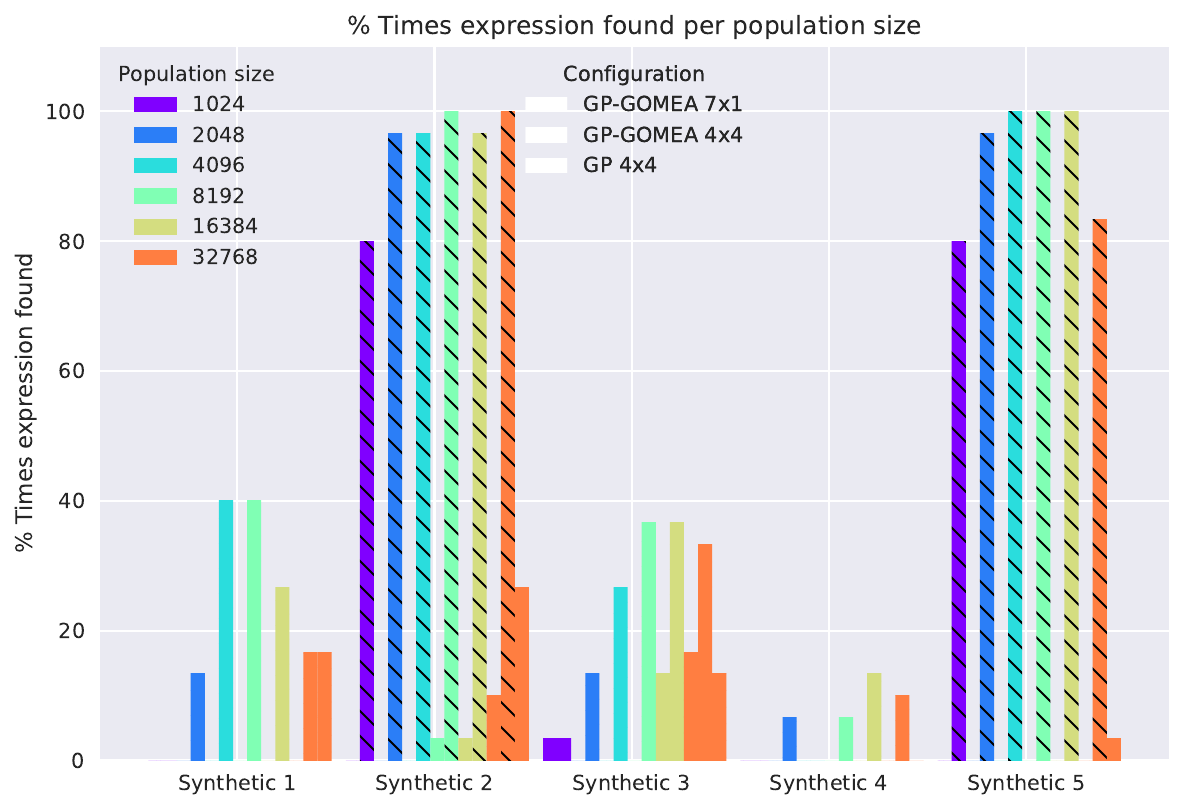}
    \caption{Comparison between the ability of different configurations of GP-GOMEA (4 tree templates of depth 4, 1 tree template of depth 7) to recover expressions.}
    \label{fig:expressionfound}
\end{figure}

In Figure \ref{fig:expressionfound} we observe that the modular configuration of GP-GOMEA is exceedingly better at recovering the synthetic expressions, with the non-modular configuration of GP-GOMEOA and the depth-constrained modular configuration of GP hardly recovering any expressions. We observe a similar pattern for the obtained $R^2$: the population size needs to be large enough, but not too large (as then it limits the number of achievable generations too much). 

There is also a striking difference in the difficulty of recovering the different expressions. Expressions 2 and 5 are frequently being found by the modular configuration of GP-GOMEA whereas expressions 1, 3 and 4, although reaching high $R^2$ in Figure \ref{fig:r2perpopsize}, are notably more difficult to recover. 

Expressions 1 and 3 share a similar structure, namely the repeated addition of the same subexpression with different inputs. A common challenge in such cases is the presence of deceptive local attractors: subexpressions that are close in performance to the optimal one can pull the search toward a suboptimal expression. For example, in the case of expression 1, the term $t_0=cos(x_1+x_5)$ is very close in performance to the term $t_1=sin(x_0+x_1)$. Potentially, $t_0$ is available in the population while $t_1$ is needed but not (readily) available. Individuals undergoing GOM can improve their fitness by using $t_0$ and adding the other required terms until it is the only term that still needs changing. At this point, $t_0$ has already spread to the entire population, and it is impossible to build term $t_1$ because the necessary building blocks are no longer available. This type of convergence toward an easier-to-discover but ultimately incorrect subexpression can significantly reduce the algorithm’s ability to escape local optima.
\subsection{Modular optimisation speed}

\begin{figure}[H]
    \centering
    \begin{subfigure}[b]{0.48\textwidth}
        \centering
        \includegraphics[width=\textwidth]{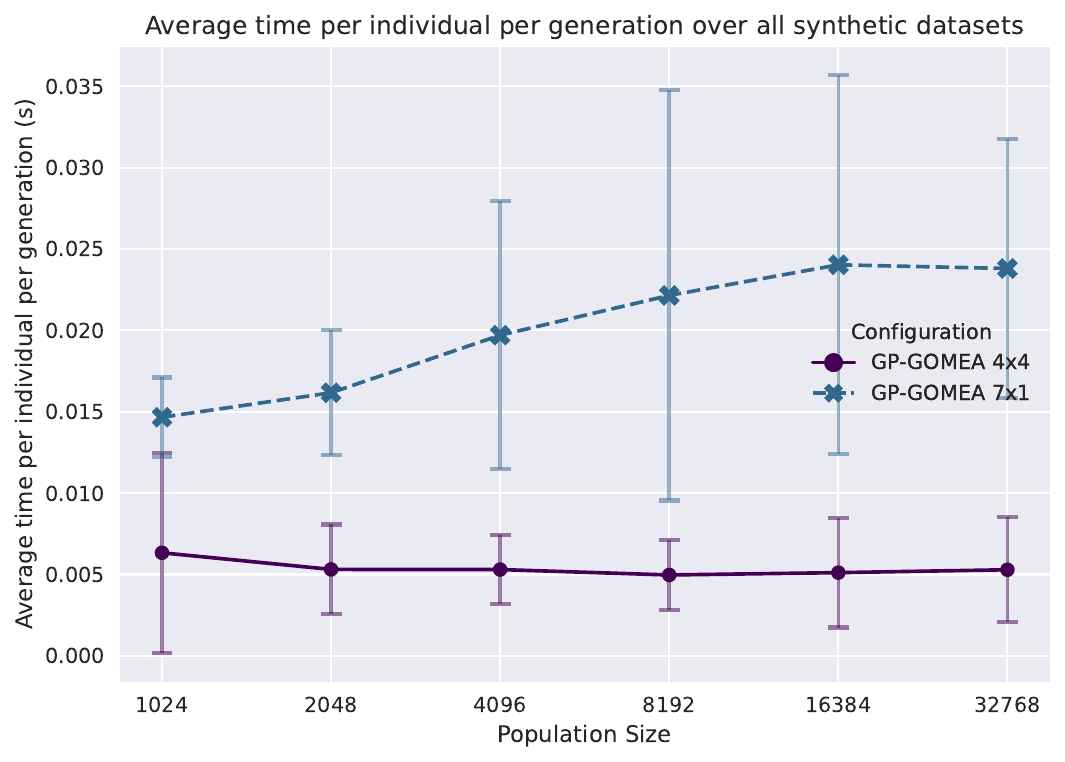}
        \caption{Average time per individual per generation per population size.}
        \label{fig:timepergenperpop}
    \end{subfigure}
    \hfill
        \begin{subfigure}[b]{0.48\textwidth}
        \centering
        \includegraphics[width=\textwidth]{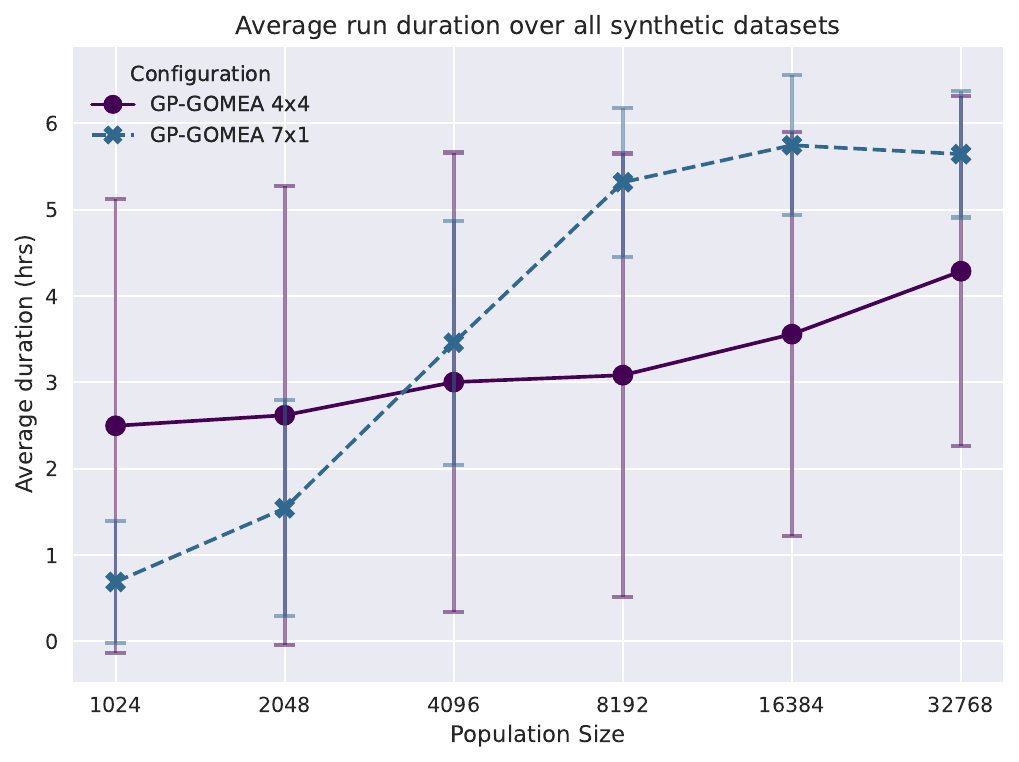}
        \caption{Average duration per population size.
        \\}
        \label{fig:durationperpop}
    \end{subfigure}

     \centering
    \begin{subfigure}[b]{0.48\textwidth}
        \centering
        \includegraphics[width=\textwidth]{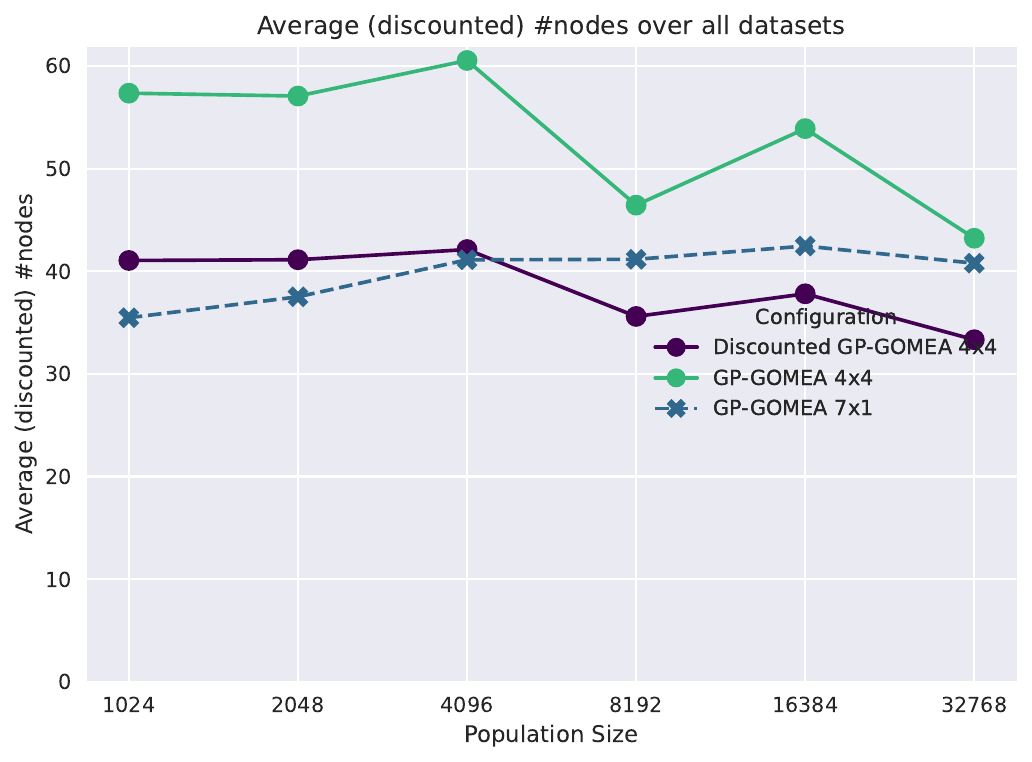}
        \caption{Average \#discounted nodes of elite per population size.}
        \label{fig:discountedperpop}
    \end{subfigure}
    \hfill
    \begin{subfigure}[b]{0.48\textwidth}
        \centering
        \includegraphics[width=\textwidth]{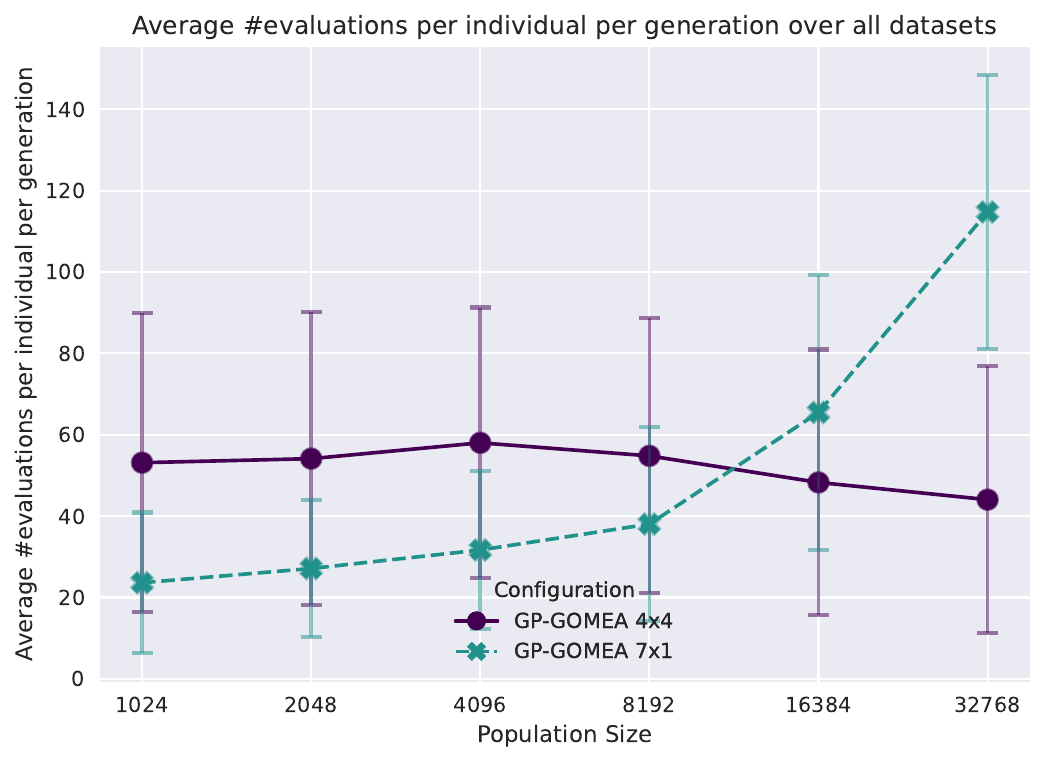}
        \caption{Average \#evaluations per population size.}
        \label{fig:evalsperpop}
    \end{subfigure}
    \caption{}
    \caption{Comparison in average time per individual per generation, average duration, average \# nodes, and the average \#evalutations per indivdiual per generation between different configurations of GP-GOMEA (4 trees of depth 4, 1 tree of depth 7). Averaging is done over all individuals and all runs of all 5 synthetic datasets. Whiskers indicate the standard deviation range.}
    \label{fig:combinedtime}
\end{figure}

In Figure \ref{fig:timepergenperpop} we observe that in the modular configuration of GP-GOMEA it takes less time per individual per generation on average than in the non-modular configuration and it takes less time for the modular configuration to find better solutions in general (see Figure \ref{fig:durationperpop}). This is partially due to the number of FOS elements. The 4x4 configuration has 243 FOS elements, whereas the 7x1 configuration has 508. However, this alone does not explain the increase in the time per population size of the 7x1 configuration. In the 4x4 configuration, many fitness evaluations can be skipped due to subtrees being disconnected from the output, avoiding fitness evaluations, whereas the 7x1 configuration has more non-intron changes to check. 

As the population size increases, the 4x4 configuration of GP-GOMEA can evolve and re-use useful subexpressions more often, whereas the 7x1 configuration of GP-GOMEA needs to evolve the same subexpression in multiple loci, leading to more non-intron checks. In Figure \ref{fig:discountedperpop} we observe that the 4x4 configuration of GP-GOMEA uses more nodes than the 7x1 configuration, but when discounting re-used nodes it becomes apparent that effectively less evaluations are needed. The number of nodes in the 7x1 configuration of GP-GOMEA on the other hand steadily increases with the population size and with it the number of evaluations (see Figure \ref{fig:evalsperpop}).

\subsection{Varying tree-depth and the number of multi-trees}
\label{sec:varheight}
To characterize the relation between the number of multi-trees and tree depth for the modular version of GP-GOMEA, we perform experiments in which we vary these settings using a population size of 4096. In addition to the synthetic datasets defined in subsection \ref{sec:synth} we also perform experiments on real-world datasets (see Table \ref{tab:realworld}). Unlike the synthetic datasets, the real-world datasets do not have a known ground-truth expression or operator set, hence, in addition to the operator set also used for the synthetic datasets, we use coefficients and coefficient mutation as done in \cite{virgolin2022coefficient} and linear scaling to further improve $R^2$ \cite{keijzer2003improving} (also used in GP-GOMEA in \cite{virgolin2021improving}). Experiments have a time budget of \textbf{1 hour}, or can terminate if the non-dominated front has not improved for 100 consecutive generations, or if $R^2$ reaches 1.

\begin{figure}[h]
    \centering
    \begin{subfigure}[b]{0.48\textwidth}
        \centering
        \includegraphics[width=\textwidth]{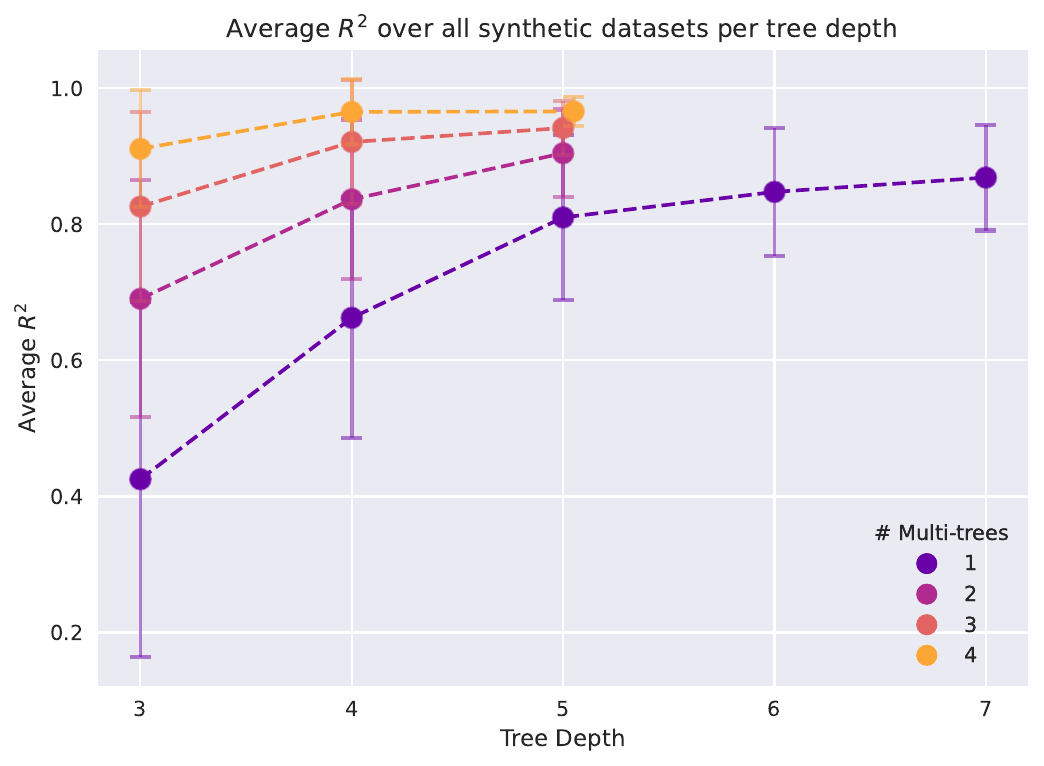}
        \caption{Results synthetic datasets}
        \label{fig:syntheticr2-1}
    \end{subfigure}
    \hfill
    \begin{subfigure}[b]{0.48\textwidth}
        \centering
        \includegraphics[width=\textwidth]{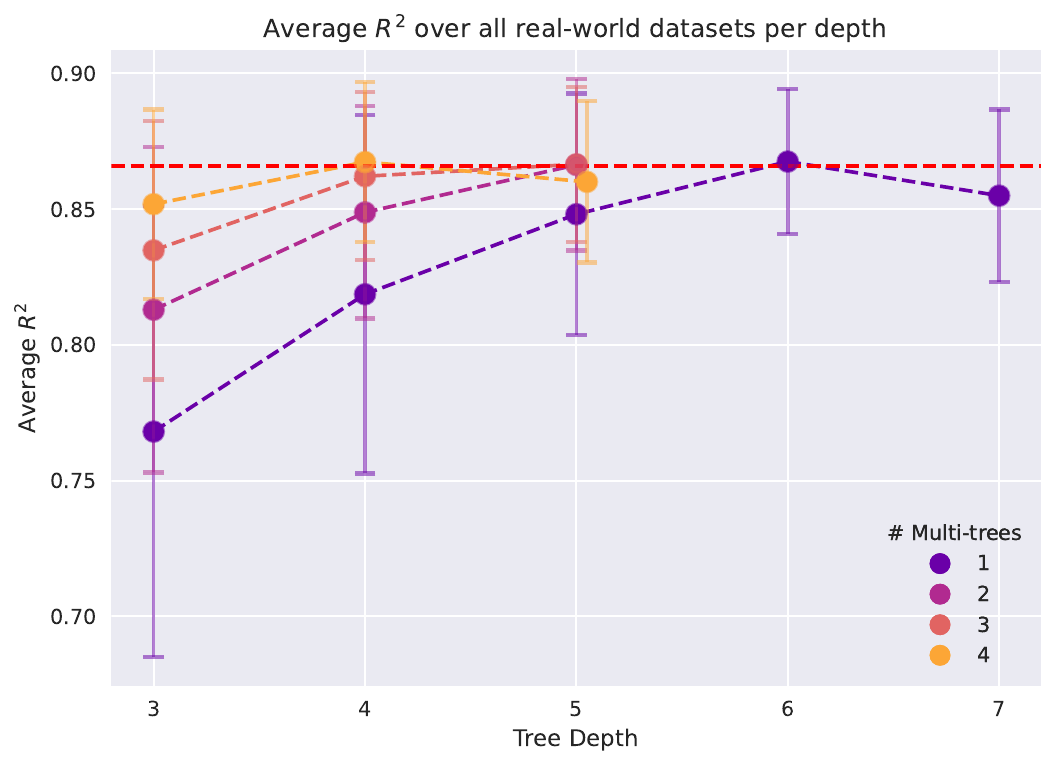}
        \caption{Results real-world datasets}
        \label{fig:realworldr2}
    \end{subfigure}
    \caption{Average $R^2$ of best individual per tree depth and per number of multi-trees. Whiskers indicate the standard deviation range.}
    \label{fig:combinedr2}
\end{figure}

\begin{figure}[H]
    \centering
    \begin{subfigure}[b]{0.48\textwidth}
        \centering
        \includegraphics[width=\textwidth]{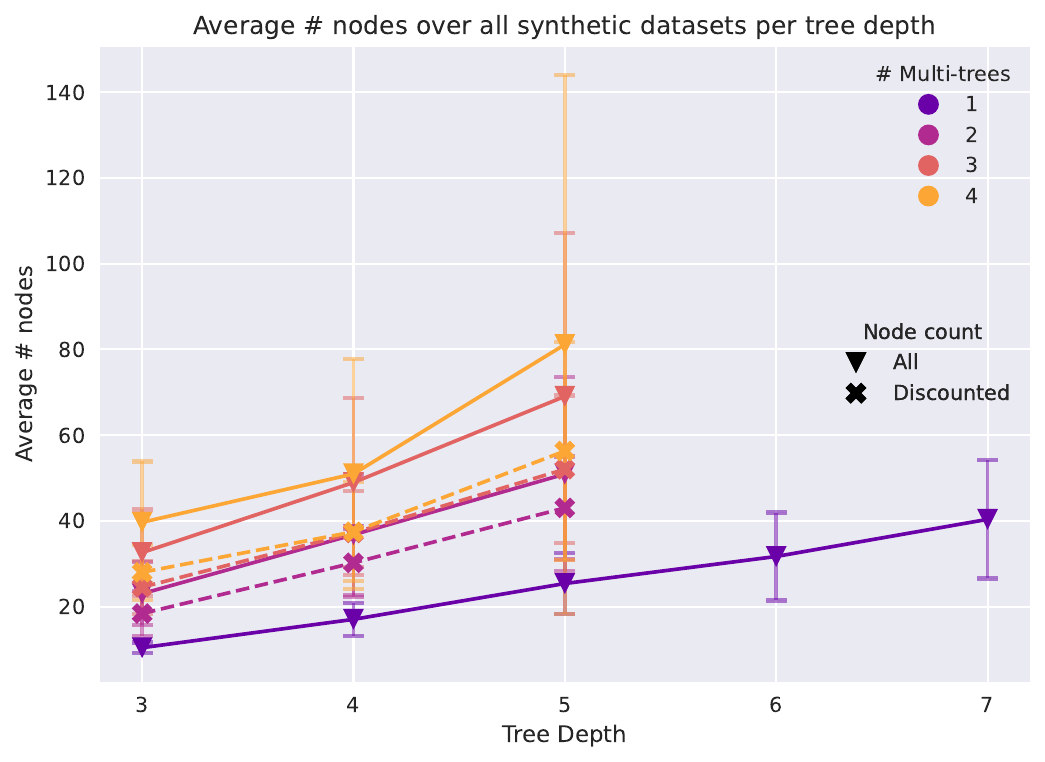}
        \caption{Results synthetic datasets}
        \label{fig:syntheticnumbernodes}
    \end{subfigure}
    \hfill
    \begin{subfigure}[b]{0.48\textwidth}
        \centering
        \includegraphics[width=\textwidth]{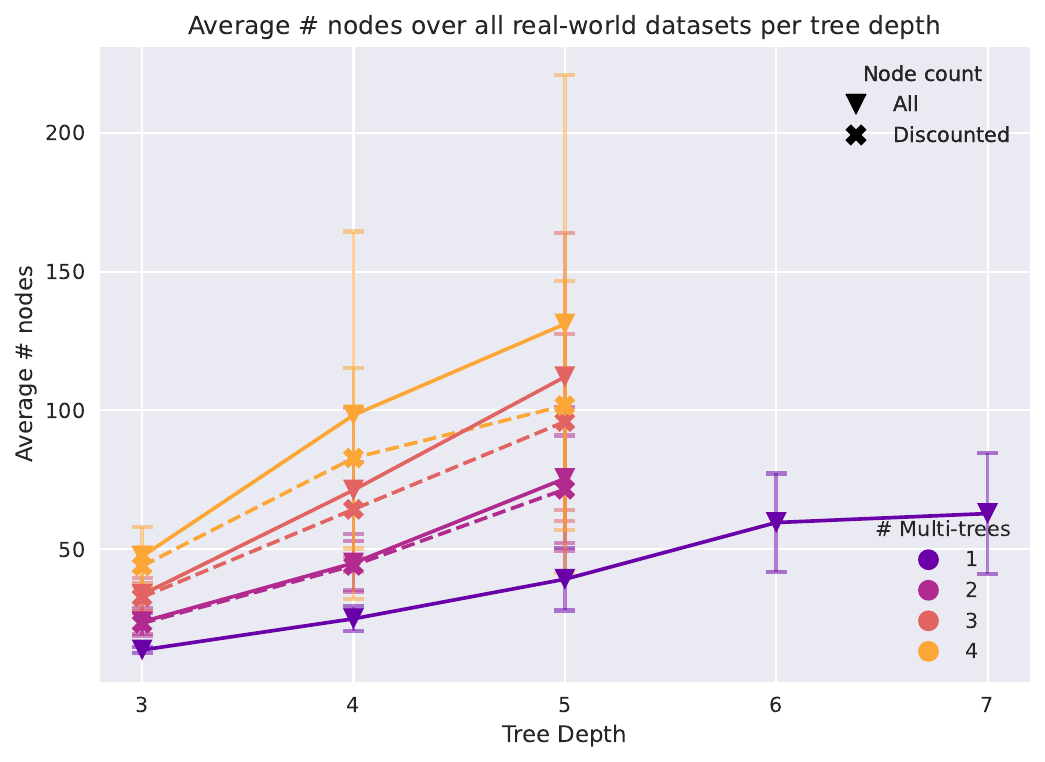}
        \caption{Results real-world datasets}
        \label{fig:realworldnumbernodes}
    \end{subfigure}
    \caption{Average \# nodes per tree depth and number of multi-trees. Whiskers indicate the standard deviation range.}
    \label{fig:combinednumbernodes}
\end{figure}

Figure \ref{fig:syntheticr2-1} shows a clear trend: increasing the number of multi-trees in GP-GOMEA and the depth of their templates, generally results in higher $R^2$ values. Only the 4x5 configuration deviates from the general trend. It performs fewer generations within the 1-hour time limit because of its larger size. As shown in Figures \ref{fig:syntheticnumbernodes} and \ref{fig:realworldnumbernodes}, this configuration also has a high number of non-intron nodes, which limits its ability to bypass costly fitness evaluations during GOM. The rate of improvement in $R^2$ diminishes with each additional multi-tree or increase in template depth. This is expected, as the number of multi-trees and their maximum depth determines the maximum size of the resulting expressions, which in turn affects accuracy. For instance, the 3x1 configuration cannot construct synthetic expression 1 due to its limited size. 

With a depth of 4, only two multi-trees are needed to reconstruct synthetic expression 1, while a single multi-tree requires a depth of 7. Comparing the average $R^2$ values of the 4x2 configuration (0.68), the 4x4 configuration (0.87), and the 7x1 configuration (0.82) on synthetic expression 1, suggests that part of the strength of Modular GP-GOMEA is in having more available structure (i.e., more multi-trees and a larger depth) rather than in the ability to re-use subexpressions as functions. While the 4x2 configuration has sufficient structure to represent the target expression, its considerably lower $R^2$ indicates that subexpression re-use alone does not explain the performance gap.

In Figure \ref{fig:realworldr2} we observe that the trend observed for the synthetic dataset again partially holds: more multi-trees and a larger depth for the tree templates generally results in better $R^2$, except for the 7x1 and 5x4 configurations, again due to early convergence. The results however are less pronounced. The gap in $R^2$ when adding multi-trees is comparatively smaller than for the synthetic datasets. The average $R^2$ of the 6x1 configuration for the real-world datasets is on par with the average $R^2$ of the 4x4 configuration. The performance difference of these configurations varies largely with the dataset. 

In Figure \ref{fig:combinednumbernodes}, we see what the effect of the available template space is on the number of nodes that are used and re-used.
In Figure \ref{fig:combinednumbernodes}, we observe that the number of nodes of resulting expressions increases with both the depth and number of multi-trees. Interestingly, even though there is no downward pressure on the number of nodes, the expressions do not use all available nodes, e.g. trees with configuration 7x1 on average use only 40 nodes out of the available 127 for the synthetic datasets. This is a direct consequence of the way in which GP-GOMEA works and the way in which templates are designed. Filling out a template completely is not trivial and moreover, since GP-GOMEA does not perform subtree swapping, bloating is harder to achieve. Still, it is easier to obtain better $R^2$ values with larger expressions, which is why ultimately GP-GOMEA does still find larger expressions than needed sometimes.

In Section \ref{sec:synth} it is mentioned that it is unknown whether the underlying expressions of real-world datasets can be decomposed. This remains unknown. In Figure \ref{fig:realworldnumbernodes} we observe from the discrepancy between the number of nodes and the number of de-duplicated nodes (i.e., the number of re-used nodes subtracted from the total number of nodes) that there is re-use in the expressions found for the real-world datasets (see also Table \ref{tab:reuse}). We also observe that the use of more trees in the representation, leads to more nodes being re-used. We hypothesize that this is partially due to the additional subexpressions being used to increase the depth of the expression. We show the average number of subexpressions and average number of re-used subexpressions for the 4x4 configuration in Table \ref{tab:reuse}. We observe that on average at least one subexpression is re-used for each real-world dataset, alluding to some real-world problems being decomposable into re-usable subexpressions. We also see that on average more subexpressions are used than re-used, indicating the need for a larger template than the standard tree template with a depth of 4 (31 nodes). We also observe that the expression evolved for the synthetic datasets does not always re-use the number of functions envisioned in the creation of the datasets. This is partially due to the expression failing to be found due to the aforementioned deceptive local attractor and partially due to the lack of parsimony pressure.

\begin{table}[h]
\centering
\begin{tabular}{lccc}
\toprule
Dataset      & Average  & Average & Average \#subexpressions\\
 & \#subexpressions used & \#subexpressions re-used & re-used as function\\
\midrule
Air          & 5.41                     & 2.76  & 0.97                       \\
Bike         & 3.70                     & 0.87  & 0.83                        \\
Concrete     & 3.67                     & 1.03  & 0.93                        \\
Dow Chemical & 3.83                     & 1.00  & 0.47                        \\
Tower        & 3.73                     & 1.03   & 0.63                         \\
Synthetic 1  & 3.30                     & 1.00  & 0.80                        \\
Synthetic 2  & 7.47                     & 4.97  & 0.13                        \\
Synthetic 3  & 6.90                     & 4.17  & 0.90                        \\
Synthetic 4  & 4.93                     & 2.10  & 0.80                        \\
Synthetic 5  & 4.03                     & 1.03  & 1.00
\\
\bottomrule              
\end{tabular}
\caption{Average use and re-use of subexpressions per dataset of best expressions for modular GP-GOMEA configuration 4x4.}
\label{tab:reuse}
\end{table}

\subsection{Insights from a qualitative perspective}
\label{sec:qual}
So far, we have seen that having the ability to evolve modular subexpressions can improve the accuracy of the overall expression quantitatively, but what do these expressions and subexpressions look like, and do they improve interpretability?

In this subsection, we randomly pick an expression with re-used subexpressions (i.e. trees in the multi-tree that are referenced more than once) from the non-dominated front resulting from running the 4x4 configuration of modular GP-GOMEA on the Bike Daily dataset and attempt to gain qualitative insights into interpretability by inspecting how an input feature impacts the value of a subexpression. To do this we fix a set of input features to random values in their range and choose a random input feature $x$ as a variable (the weather situation in Expression 1). The Bike Daily dataset is chosen because it does not require deep expert knowledge about the data. Further information regarding the dataset's features can be found in \cite{asuncion2007uci}. For clarity, expressions have been simplified, and real values have been truncated to two decimals.

\begin{align*}
    \text{Expression}_1 = & \ 3414.75 + 3414.75 \times f_0 \times \sin(2 \times \text{Temperature}) \\
    & + f_0 \times \sin(2.62) \\
    & - (\text{Humidity} + \text{Holiday}) \times \text{Humidity} \\
    f_0 = & \ \sin\left(\frac{\text{Season}}{4.97}\right) - \cos\left(\frac{\text{Temperature}}{-0.19}\right) \\
    & - \cos(\text{Holiday} \times \text{Temperature}) - \sin(\sin(x)) \\
    & + \left|\frac{-0.64}{\text{Windspeed} \times \cos(\text{Humidity})}\right| \\
    & - \sqrt{\frac{\text{Year} \times \cos(x)}{x}}
\end{align*}

\begin{figure}
\centering
\begin{subfigure}[H]{.5\textwidth}
  \centering
  \includegraphics[width=\linewidth]{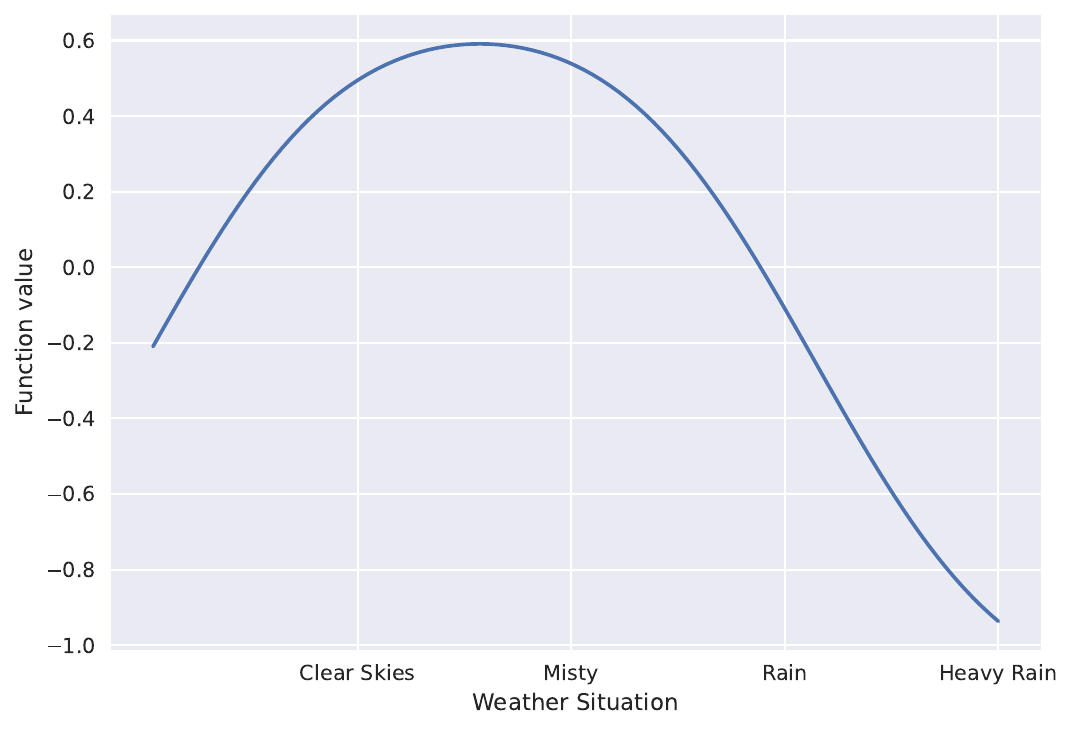}
  \caption{Values set: Season=1, Temperature=0.34 (normalised), Windspeed=0.16, Humidity (normalised)=0.81}
  \label{fig:sub1}
\end{subfigure}%
\begin{subfigure}[H]{.5\textwidth}
  \centering
  \includegraphics[width=\linewidth]{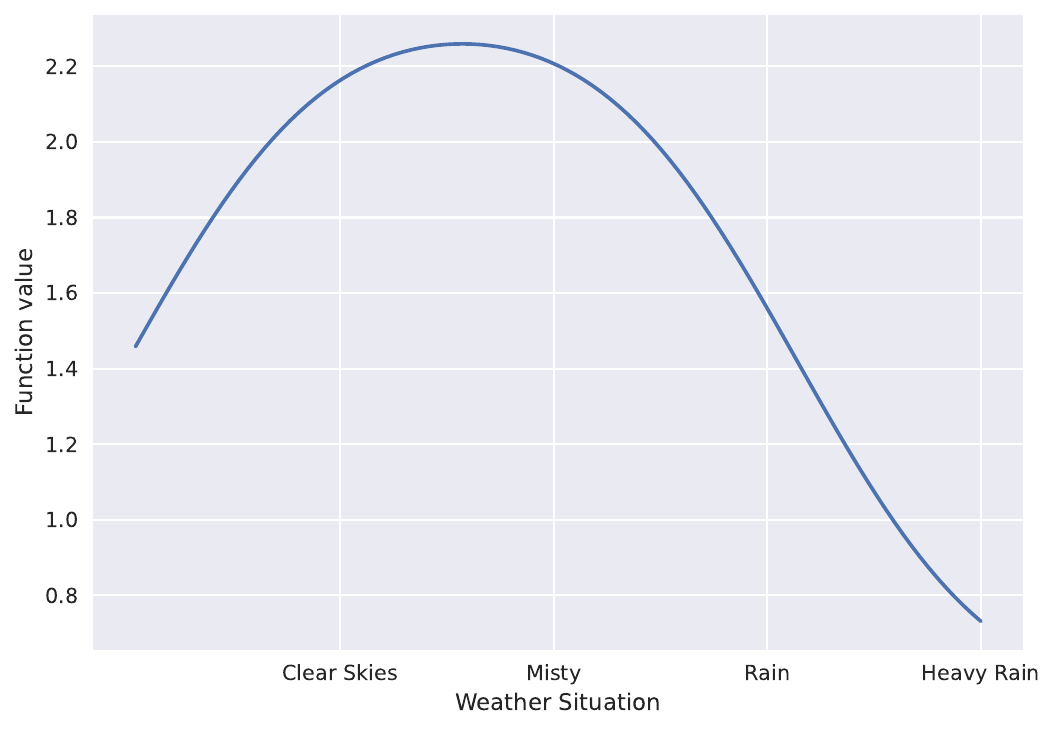}
  \caption{Values set: Season=2, Temperature=0.66 (normalised), Windspeed=0.51, Humidity (normalised)=0.39}
  \label{fig:sub2}
\end{subfigure}
\caption{Plots of subexpression obtained with Modular GP-GOMEA. The randomly set variables are mentioned in the captions of the subplots.}
\label{fig:test}
\end{figure}

We observe in Figure \ref{fig:sub1} that the subexpression yields a higher value in the absence of rain (clear skies, misty) and a high value during (heavy) rainfall. Similarly, in Figure \ref{fig:sub2}, with different values set, the general shape of the subexpression stays the same and only the function value is offset. Interpreting how these values fit into the entire expression is more challenging, but intuitively, this subexpression can be employed as a feature that correlates well with the number of bikes rented. Understanding the subexpression simplifies the entire expression, as it only needs to be understood once and can then be applied wherever it is re-used. If the second occurrence of $f_0$ were a very different-looking subexpression that essentially does something very similar, something that may well happen in symbolic regression, it would take much longer to understand the expression.

Note that the variable for this subexpression is categorical and non-ordinal, yet a meaningful subexpression can still be constructed. Also interesting to note is that the term $sin(2.62)$ could be simplified to a single coefficient, but due to the lack of parsimony pressure, it occurs in the expression in this form.

\section{Discussion}
This work introduces modularity into GP-GOMEA by means of a multi-tree template representation in which some trees can call other trees. This representation, significantly increases the diversity of tree shapes that can be efficiently evolved. While the algorithm still relies on fixed-sized templates, which imposes some structural constraints, this limitation can actually be seen as beneficial. By bounding the size of each (sub)expression, the risk of excessive bloat \cite{langdon1998fitness} is reduced. In this way, the fixed-sized template supports the creation of more manageable, interpretable expressions and subexpressions, balancing flexibility in tree structure with interpretability.

The interpretability of symbolic expressions is influenced by their size. Intuitively, the re-use of subexpressions should reduce the cognitive load required to understand these expressions, as smaller, repeated components are easier to grasp than large, complex ones. However, no empirical evidence through user studies currently supports this intuition in the context of symbolic SR. Further studies or surveys on the impact of subexpression re-use on interpretability are necessary before any definitive conclusions can be drawn. In any case, Modular GP-GOMEA is an excellent algorithm for discovering such expressions while maintaining competitive training errors.

Some experiments were conducted on synthetic data based on target expressions that are known to contain subexpressions. Modular GP-GOMEA was found to be able to achieve what it was designed for and boasts a clear advantage over single-template GP-GOMEA due to its ability to exploit subexpression re-use. Modular GP-GOMEA performs on par with standard GP-GOMEA on real-world datasets but the latter uses a much larger structure. In most real-world cases, subexpressions are used to leverage more flexible structures rather than being re-used, a phenomenon that also unexpectedly occurs in synthetic datasets. Introducing parsimony pressure, e.g., through multi-objective optimization where one objective is expression size or subexpression re-use, or both, could further encourage to re-use of subexpressions. Nonetheless, even when only optimizing for accuracy Modular GP-GOMEA does find instances of re-used subexpressions, one of which is inspected in Section \ref{sec:qual} and was found to make qualitative sense by the authors, but also still exhibits superfluous constructs that could be simplified further. 

Regarding the experiments, due to time constraints, only a limited number of population sizes were tested in this work. To avoid introducing another variable in Section \ref{sec:varheight}, the population size used is the best-tested one for 2 out of 5 synthetic problems and was not tested for real-world problems. It is possible, and even likely, that each specific problem requires population size tuning to achieve optimal performance. However, we believe that demonstrating the relationship between depth, the number of multi-trees, and $R^2$ is sufficient with just one population size. Moreover, the interleaved multi-start scheme was introduced also for GP-GOMEA \cite{virgolin2021improving}. It could similarly be used for modular GP-GOMEA, removing the need to set a population size in practice.

A limitation of our experiments is the use of a fixed time budget of 6 hours or 1 hour. This choice may benefit certain problems and algorithms; for example, the time required for the modular configuration on problem 2 is much shorter than that for the non-modular configuration on problem 1. Running the experiments for longer periods is likely to yield better results, and a time-to-optimum comparison may provide a more comprehensive evaluation. However, we consider such experiments to be out of scope for this work, which focuses on the introduction of modular GP-GOMEA and the identification of its potential.
\section{Conclusions}
In this paper, we introduced \textbf{Modular GP-GOMEA}, an extension of GP-GOMEA that uses a multi-tree template, allowing trees to reference other trees as subexpressions. This representation allows for the construction of more expressive and complex symbolic expressions without significantly compromising performance. We also developed a set of synthetic benchmark expressions that emphasize the need for functional re-use. Modular GP-GOMEA can efficiently recover these types of expressions at a faster rate compared to the original GP-GOMEA and template-constrained GP.

Moreover, Modular GP-GOMEA achieved higher $R^2$ across five real-world datasets, clearly indicating an improvement in accuracy. Increasing the number of multi-trees and the depth of each tree generally enhances $R^2$, provided there is enough time budget; however, caution is necessary to avoid excessive complexity to occur in expressions, as this can hinder interpretability.

Our analysis reveals that while Modular GP-GOMEA primarily benefits from its flexible and expansive template structure, it utilizes subexpression re-use in the form of functions less frequently than anticipated. Future work should explore the implementation of parsimony pressure to encourage greater re-use of functional subexpressions. Nonetheless, the potential for re-using subexpressions to reduce the cognitive load required to interpret full expressions is promising, warranting further investigation.

In conclusion, Modular GP-GOMEA marks a significant advancement over standard GP-GOMEA, providing more flexible templates and the potential for generating interpretable expressions through automatic subexpression discovery. While further improvements are possible, including multi-objectivization and the use of more advanced ways to perform GOM and the use of higher-arity tree templates \cite{schlender2024improving}, the concept of modular GP-GOMEA was shown to hold great promise both from an interpretability point of view as well as in terms of enhanced capabilities to discover highly accurate symbolic expressions.
\subsection{Acknowledgements}
This research is part of the research programme Open Competition Domain Science-KLEIN with project number OCENW.KLEIN.111, which is financed by the Dutch Research Council (NWO). We further thank the Maurits en Anna de Kock Foundation for financing a high-performance computing system.
\bibliographystyle{splncs04}
\bibliography{refs}

\begin{thebibliography}{10}
\providecommand{\url}[1]{\texttt{#1}}
\providecommand{\urlprefix}{URL }
\providecommand{\doi}[1]{https://doi.org/#1}

\bibitem{asuncion2007uci}
Asuncion, A., Newman, D.: Uci machine learning repository (2007)

\bibitem{doshi2017towards}
Doshi-Velez, F., Kim, B.: Towards a rigorous science of interpretable machine learning. arXiv preprint arXiv:1702.08608  (2017)

\bibitem{gerules2016survey}
Gerules, G., Janikow, C.: A survey of modularity in genetic programming. In: 2016 IEEE Congress on Evolutionary Computation (CEC). pp. 5034--5043. IEEE (2016)

\bibitem{goodman2017european}
Goodman, B., Flaxman, S.: European union regulations on algorithmic decision-making and a “right to explanation”. AI magazine  \textbf{38}(3),  50--57 (2017)

\bibitem{gronau2007optimal}
Gronau, I., Moran, S.: Optimal implementations of upgma and other common clustering algorithms. Information processing letters  \textbf{104}(6),  205--210 (2007)

\bibitem{harrison2022gene}
Harrison, J., Alderliesten, T., Bosman, P.A.: Gene-pool optimal mixing in cartesian genetic programming. In: International Conference on Parallel Problem Solving from Nature. pp. 19--32. Springer (2022)

\bibitem{keijzer2003improving}
Keijzer, M.: Improving symbolic regression with interval arithmetic and linear scaling. In: European Conference on Genetic Programming. pp. 70--82. Springer (2003)

\bibitem{kommenda2018local}
Kommenda, M., Affenzeller, M.: Local optimization and complexity control for symbolic regression. Ph. D. dissertation, Ph. D. thesis  (2018)

\bibitem{kommenda2020parameter}
Kommenda, M., Burlacu, B., Kronberger, G., Affenzeller, M.: Parameter identification for symbolic regression using nonlinear least squares. Genetic Programming and Evolvable Machines  \textbf{21}(3),  471--501 (2020)

\bibitem{koza1993hierarchical}
Koza, J.R.: Hierarchical automatic function definition in genetic programming. In: Foundations of Genetic Algorithms, vol.~2, pp. 297--318. Elsevier (1993)

\bibitem{koza1994genetic}
Koza, J.R.: Genetic programming as a means for programming computers by natural selection. Statistics and computing  \textbf{4},  87--112 (1994)

\bibitem{koza1994scalable}
Koza, J.R., Kinner, K., et~al.: Scalable learning in genetic programming using automatic function definition. Advances in Genetic Programming pp. 99--117 (1994)

\bibitem{la2021contemporary}
La~Cava, W., Orzechowski, P., Burlacu, B., de~Fran{\c{c}}a, F.O., Virgolin, M., Jin, Y., Kommenda, M., Moore, J.H.: Contemporary symbolic regression methods and their relative performance. arXiv preprint arXiv:2107.14351  (2021)

\bibitem{langdon1998fitness}
Langdon, W.B., Poli, R.: Fitness causes bloat. Springer (1998)

\bibitem{lipton2018mythos}
Lipton, Z.C.: The mythos of model interpretability: In machine learning, the concept of interpretability is both important and slippery. Queue  \textbf{16}(3),  31--57 (2018)

\bibitem{luong2012elitist}
Luong, H.N., Bosman, P.A.: Elitist archiving for multi-objective evolutionary algorithms: To adapt or not to adapt. In: Parallel Problem Solving from Nature-PPSN XII: 12th International Conference, Taormina, Italy, September 1-5, 2012, Proceedings, Part II 12. pp. 72--81. Springer (2012)

\bibitem{mei2022explainable}
Mei, Y., Chen, Q., Lensen, A., Xue, B., Zhang, M.: Explainable artificial intelligence by genetic programming: A survey. IEEE Transactions on Evolutionary Computation  \textbf{27}(3),  621--641 (2022)

\bibitem{ribeiro2016should}
Ribeiro, M.T., Singh, S., Guestrin, C.: " why should i trust you?" explaining the predictions of any classifier. In: Proceedings of the 22nd ACM SIGKDD international conference on knowledge discovery and data mining. pp. 1135--1144 (2016)

\bibitem{samek2017explainable}
Samek, W.: Explainable artificial intelligence: Understanding, visualizing and interpreting deep learning models. arXiv preprint arXiv:1708.08296  (2017)

\bibitem{schlender2024improving}
Schlender, T., Malafaia, M., Alderliesten, T., Bosman, P.A.: Improving the efficiency of gp-gomea for higher-arity operators. arXiv preprint arXiv:2402.09854  (2024)

\bibitem{schmidt2009distilling}
Schmidt, M., Lipson, H.: Distilling free-form natural laws from experimental data. science  \textbf{324}(5923),  81--85 (2009)

\bibitem{thierens2011optimal}
Thierens, D., Bosman, P.A.: Optimal mixing evolutionary algorithms. In: Proceedings of the 13th annual conference on Genetic and evolutionary computation. pp. 617--624 (2011)

\bibitem{virgolin2017scalable}
Virgolin, M., Alderliesten, T., Witteveen, C., Bosman, P.A.: Scalable genetic programming by gene-pool optimal mixing and input-space entropy-based building-block learning. In: Proceedings of the Genetic and Evolutionary Computation Conference. pp. 1041--1048 (2017)

\bibitem{virgolin2021improving}
Virgolin, M., Alderliesten, T., Witteveen, C., Bosman, P.A.: Improving model-based genetic programming for symbolic regression of small expressions. Evolutionary computation  \textbf{29}(2),  211--237 (2021)

\bibitem{virgolin2022coefficient}
Virgolin, M., Bosman, P.A.: Coefficient mutation in the gene-pool optimal mixing evolutionary algorithm for symbolic regression. In: Proceedings of the Genetic and Evolutionary Computation Conference Companion. pp. 2289--2297 (2022)

\bibitem{walker2008automatic}
Walker, J.A., Miller, J.F.: The automatic acquisition, evolution and reuse of modules in cartesian genetic programming. IEEE Transactions on Evolutionary Computation  \textbf{12}(4),  397--417 (2008)

\end{thebibliography}
\clearpage

%
%
%

%
\appendix
\section{Adding extra inputs}
\label{sec:inputs}
In the original formulation by Koza, HADFs only accept arguments from the main tree as terminals. The terminal set for the subexpression thus only has argument nodes, whereas in Modular GP-GOMEA, ERCs and input features are included.
\begin{figure}[h]
    \centering
    \includegraphics[width=0.7\textwidth]{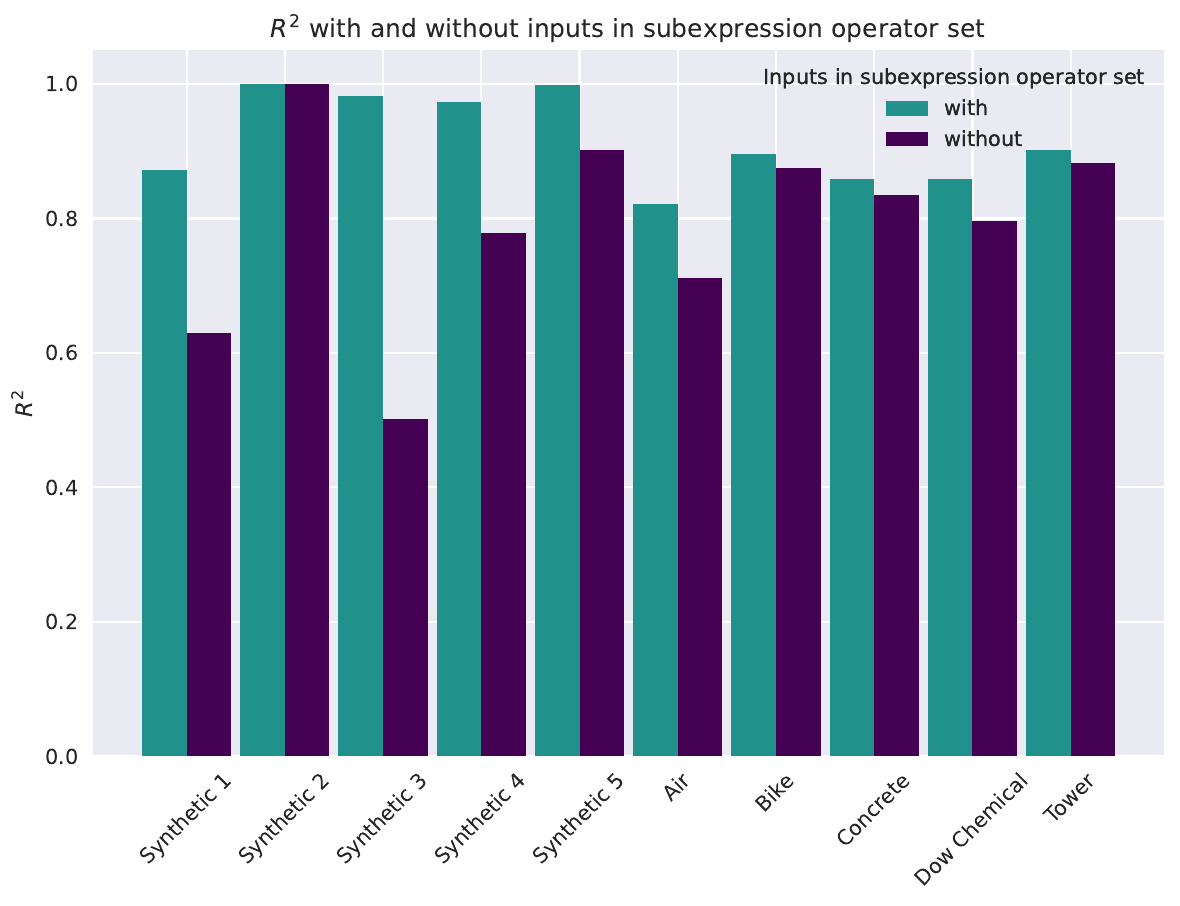}
    \caption{$R^2$ with feature inputs in subexpression operator set and without (Koza's formulation).}
    \label{fig:drift}
\end{figure}
\end{document}